\pdfoutput=1

\documentclass[11pt]{article}

\usepackage[]{EMNLP2023}

\usepackage{times}
\usepackage{latexsym}
\usepackage{floatrow}

\usepackage[T1]{fontenc}
\usepackage[utf8]{inputenc}
\usepackage{microtype}
\usepackage{float}
\usepackage{inconsolata}
\usepackage{tabularx}
\usepackage{comment}
\usepackage{graphicx}
\usepackage{subcaption}
\usepackage{booktabs}
\usepackage{multirow}
\usepackage{adjustbox}
\usepackage{todonotes}
\usepackage{amsmath}

\newcommand{\ggrev}[2]{\textcolor{red}{#2}}
\title{Quantifying Divergence for Human-AI Collaboration and Cognitive Trust}

\author{
\centering
\begin{minipage}[t]{\textwidth}
\centering
\normalsize
\bf
Müge Kural,$^{1,2,3}$ Ali Gebeşçe,$^{1,3}$ Tilek Chubakov,$^{1,3*}$
Gözde Gül Şahin$^{1,2,3}$\\
{\footnotesize \normalfont
$^{1}$ Koç University
Computer Science and Engineering
Department \
$^{2}$  KUIS AI \
$^{3}$  \
\url{https://gglab-ku.github.io/}
}
\end{minipage}
}

\begin{document}
\maketitle
\begin{abstract}

Predicting the collaboration likelihood and measuring cognitive trust to AI systems is more important than ever. To do that, previous research mostly focus solely on the model features (e.g., accuracy, confidence) and ignore the human factor. To address that, we propose several \textit{decision-making similarity} measures based on divergence metrics (e.g., KL, JSD) calculated over the labels acquired from humans and a wide range of models. We conduct a user study on a textual entailment task, where the users are provided with soft labels from various models and asked to pick the closest option to them. The users are then shown the similarities/differences to their most similar model and are surveyed for their likelihood of collaboration and cognitive trust to the selected system. Finally, we qualitatively and quantitatively analyze the relation between the proposed \textit{decision-making similarity} measures and the survey results.
 
We find that people tend to collaborate with their most similar models---measured via JSD---yet this collaboration does not necessarily imply a similar level of cognitive trust. We release all resources related to the user study (e.g., design, outputs), models, and metrics at our repo\footnote{https://github.com/gglab-ku/cogeval}. 

\end{abstract}

\section{Introduction}
\begin{figure}[ht!]
 \begin{flushright}
     \includegraphics[scale=0.25]{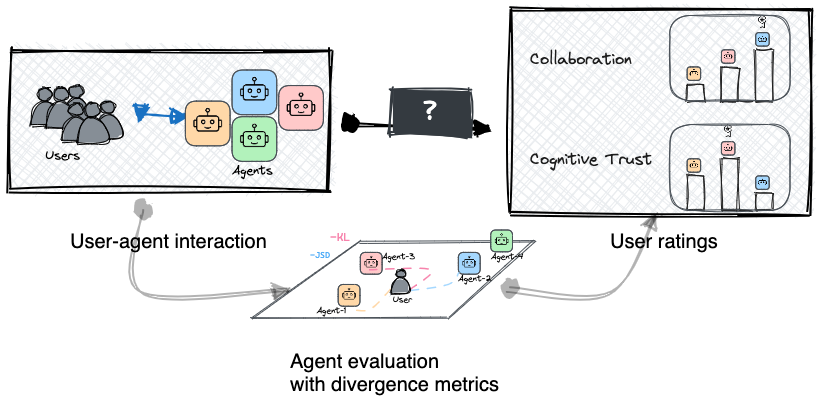}

 \end{flushright}
    \caption{Decision-making similarities between the user and various models are calculated using various divergence metrics, then linked to collaboration preferences and cognitive trust.}
  \label{fig:main-figure}
\end{figure}

Human-AI interaction is becoming increasingly important in our daily lives as the number of AI assistants such as ChatGPT~\cite{openai_2022} grows. Humans employ these assistants in many different areas ranging from healthcare~\cite{jeblick2023chatgpt, liu2023utility, howard2023chatgpt} to education~\cite{moore2022assessing, abdelghani2023gpt, dijkstra2022reading}. In a survey conducted by \citet{skjuve4376834people}, authors find that individuals use ChatGPT with the motivations such as productivity, novelty, creative work, learning and development. 
For instance, developers use these assistants to increase their productivity. \citet{barke2023grounded} highlight that developers use them \footnote{https://github.com/features/copilot} both to expedite their coding processes and explore novel coding approaches. 
People also use these assistants for producing creative or formal content, e.g., writing articles, stories or formal letters. 
Finally, they are more commonly utilized as knowledge repositories akin to search engines, e.g., how an air conditioner cools a room, to enhance the knowledge of their users. 
This increasing amount of interaction raises two critical questions: How much do we prefer to collaborate with AI models where we seek their assistance, and how much do we trust them? Predicting and measuring these before deploying AI models in real-life scenarios is crucial to foresee how the interaction will unfold. Therefore, understanding the collaboration and cognitive trust dynamics is crucial to build future AI assistants.

Previous works predominantly focus on predicting the likelihood of collaboration using model-only or user-only features. For instance, \citet{vodrahalli2022uncalibrated} develop a human behavior model to predict when users will accept AI recommendations. The study incorporates a range of features, including model confidence, user confidence, demographic information (e.g., age, sex, and education level), and user opinions for AI  (e.g., how often AI aids in user's daily tasks). \citet{yin2019understanding} examine the impact of the stated and observed accuracy of the models during human-AI interactions on the trust of the users. In addition to common model features like accuracy, the effect of model generated explanations on human collaboration and trust~\cite{bansal2021does,kocielnik2019will,SHIN2021102551,zhang2020effect} are explored. On a similar line, \citet{bansal2019beyond}, highlight the importance of model predictability, showing that when participants \textit{know} when AI would succeed/fail, the level of collaboration increases. In short, existing work either employ shallow (e.g., accuracy), or sophisticated (e.g., explanations, predictability) model features; along with shallow user features~(e.g., age). 
However, studying user and model features \textit{jointly} to \textit{quantify} and \textit{predict} the level of collaboration and cognitive trust to AI, is left unexplored to the best of our knowledge.

To address that, we propose a novel experimental setup sketched in Fig.~\ref{fig:main-figure} to analyze the \textit{decision-making similarity (DMS)} between human and AI models and its impact on collaboration and cognitive trust. We first implement a diverse set of models ranging from statistical models to GPT-family ones specifically: tf-idf, Enhanced LSTM \cite{chen2016enhanced}, RoBERTa \cite{Liu2019RoBERTaAR}, and davinci-003 \cite{brown2020language} on Natural Language Inference~(NLI) task. 
Then we conduct a multi-staged user study: In the first stage, we present the generated label distributions, a.k.a. \textit{soft-labels}, from the models to participants and ask them to choose the closest label distribution; where in the second stage, users answer various questions regarding collaboration and cognitive trust to the \textit{closest} model. We define the distance as the divergence between the human-AI predictions, i.e., soft labels, using various metrics (e.g., Jensen-Shannon Distance (JSD)~\cite{Lin1991DivergenceMB}, Kullback-Leibler (KL)~\cite{kullback1951information}). Finally we analyze the relation between these metrics, a.k.a., DMS, and the user ratings for collaboration and cognitive trust.    

Our initial findings suggest that: (1) People tend to collaborate with AI models when their decision-making processes---measured via DMS---are similar. However, they are less likely to develop cognitive trust in these models. \ggrev{Model adı versek daha güzel olur. Iki durum da davinci oldugundan buraya eklemiyorum.}{}Surprisingly, they may exhibit stronger collaboration tendencies with AI models even when their trust in the model is low. \ggrev{such as Davinci}{} (2) A particular type of DMS, low inverse KL divergence, has the most influence on collaboration likelihood: agreeing on the same answer (right or wrong) with high confidence. (3) Finally, cognitive trust seems also related to low inverse-KL; however, it might also require another type of DMS with low forward-KL, i.e., avoiding being overconfident in case of uncertainty. Therefore, its likelihood can be indicated by JSD. It should be noted that, our findings are indeed encouraging, however, are bound to the small set of experimented models, task, dataset and user pool. With this work, we hope to pave the way towards incorporating the soft-labels to user studies and use the suggested divergence metrics to approximate collaboration and trust to AI models \textit{before model deployment}.

Our contributions can be listed below:
\begin{itemize}
    
    \item To the best of our knowledge, this is the first study that investigates the impact of ``decision-making similarity'' (DMS)---measurable via divergence metrics on soft labels---on human-AI collaboration and cognitive trust AI. 
    \item We propose a comprehensive, yet, flexible four-staged user study to measure DMS. Although originally designed for the NLI task, it is easily adaptable to any classification task.
    \item To encourage further studies on collaboration and trust, we share all resources, including the user study design, participant outputs, NLI models and predictions and related implementation.
\end{itemize}

\begin{table*}[!htp]
\centering
\resizebox{\textwidth}{!}{\begin{tabular}{lllcl}
\toprule
 \textbf{Premise} & \textbf{Hypothesis} & \textbf{Soft Labels} & \textbf{Hard Label} & \textbf{Type} \\
 \midrule
Two kids playing on a street. & \textbf{I1:} Two children have fun on a street. & E, E, E, N, N & E & 3GS \\
A group of bikers pedal along. & \textbf{I2:} A group of cyclists are in a marathon. & N, N, N, N, E & N & 4GS \\
A man fishing in the ocean. & \textbf{I3:} A man is driving a car. & C, C, C, C, C & C & 5GS \\
\bottomrule
\end{tabular}
}
\caption{
\label{citation-guide}
A sample of validated SNLI Pairs from \citep{bowman-etal-2015-large}. E: Entailment, N: Neutral, C: Contradiction}
\label{table: sample_snli_pairs}
\end{table*}

\section{Task Setup}
\label{sec:task_setup}
To link the dimensions of human-AI interaction with the model features, we first identify a downstream task that requires reasoning (see \S\ref{ssec:dataset}).
Then we train a diverse set of models and acquire soft labels (see \S\ref{ssec:models}). Finally, we propose and calculate several metrics to calculate the distance between the labels provided by humans and models (see \S\ref{sec:similarity}), which we refer to as \textit{decision-making similarity}.

\subsection{Dataset}
\label{ssec:dataset}
The SNLI (Stanford Natural Language Inference) dataset \cite{bowman-etal-2015-large} is a well-known resource used to evaluate the natural language inference capabilities of models. The task is mostly associated with reasoning abilities and is part of the popular language understanding benchmark~\cite{WangPNSMHLB19}. Here, each instance is annotated by five workers, and the resulting premise-hypothesis pairs were labeled as 3GS, 4GS, or 5GS, as shown in \autoref{table: sample_snli_pairs}, where the number indicates the maximum number of agreements. To reveal differences between models and create variation in the user study, we use 4GS pairs instead of the more trivial 5GS. Finally, we reduce the sample size from 90 to 50, considering the human attention span.~\footnote{Participants must carefully examine each model's confidence score and conduct analyses on similarities and differences. The annotation of 50 instances takes around 60 minutes.} 

\subsection{Models}
\label{ssec:models}
To diversify the model features---model \textit{accuracy}, \textit{confidence}, and \textit{soft-label distribution}--- we consider both neural and non-neural models, different architectures (e.g., autoregressive decoder and bidirectional encoder), and different techniques such as fine-tuning and zero-shot prompting. We particularly include models with the highest accuracy~\cite{Liu2019RoBERTaAR,chen2016enhanced}, highest confidence~\cite{brown2020language}, diverse label distribution and include a random baseline for comparison. All the models are trained on SNLI training split by \textit{excluding} the instances that are demonstrated to human annotators.

\paragraph{Random baseline} It assigns probabilities, a.k.a., confidence scores, randomly to each label (i.e., neutral, contradiction, entailment) drawn from a uniform distribution. A small Gaussian noise with a mean of 0 and variance of 0.05 is added to the soft labels~\footnote{We use soft label, confidence scores, and probabilities interchangeably throughout the paper.} and then normalized for the user study to avoid easy recognition.

\paragraph{TF-IDF} 

We extract TF-IDF weighted lexical features from both premise and hypothesis sentences and train a statistical linear classifier using the LBFGS optimizer.

\paragraph{RoBERTa~\cite{Liu2019RoBERTaAR}} We fine-tune the \texttt{roberta-large} for pair-wise classification on SNLI training dataset by excluding the instances that are demonstrated to human annotators. 
We use the default parameters provided with AllenNLP~  \footnote{\url{https://github.com/allenai/allennlp-models/blob/main/allennlp_models/modelcards/pair-classification-roberta-snli.json}}. 

\paragraph{Enhanced LSTM~\cite{chen2016enhanced}} The model uses a chain of LSTMs. The first layer, a bi-LSTM, encodes the premise and hypothesis pairs. The model then applies a form of soft alignment between premise and hypothesis token pairs to enhance the representations. 
Finally, the representations are fed into another bi-LSTM, and then passed through a pooling operation, followed by a softmax. We train the model from scratch with the default parameters.

\begin{table}[!htbp]
    \centering
    \begin{tabular}{|l|}
        \hline
         \textit{<premise\_sentence>} \\
         Can we infer that \textit{<hypothesis\_sentence>}? \\
         Answer as yes, no, or maybe. Answer: \\
         \hline
    \end{tabular}
    \caption{Prompt template for \texttt{text-davinci-003}.}
    \label{tab:davinci_prompt}
\end{table}
\paragraph{da-vinci-003~\cite{brown2020language}} We do zero-shot prompting with \texttt{text-davinci-003} model through its publicly available API and set \texttt{max\_token} to 1. We use the prompt template similar to \citet{webson2022promptbased} given in Table~\ref{tab:davinci_prompt}. We then map the model's responses of ``yes'', ``maybe'' and ``no'' to the entailment, neutral, and contradiction labels, respectively.

\section{Similarity Calculation}
\label{sec:similarity}
We define the \textit{decision-making similarity} as the difference between two probability distributions~\footnote{We have also investigated various distance metrics mostly based on hard labels, such as minimum square error, agreement percentage and Pearson's correlation. Due to the unpromising results and the lack of space in the paper, those are ommitted.}, P and Q, measured by KL and Jensen-Shannon divergence. Here, P refers to the soft-label distributions from human annotations, whereas Q refers to the label probabilities from models, a.k.a. agents. 

\paragraph{Forward and Inverse KL Divergence} KL divergence, denoted as $D_{\text{KL}}(P \parallel Q)$ quantifies the information loss when using distribution Q instead of P:
\[ \alpha\text{-}KL = D_{\text{KL}}(P \parallel Q) = \sum_{x \in \mathcal{X}} P(x) \log \left(\frac{P(x)}{Q(x)}\right)\]
We refer to the above divergence as the \textit{forward-KL} and denote it with $\alpha$-KL. 
One would expect a high $\alpha$-KL, when human annotations are more distributed, i.e., P is wider, but the model is overconfident on one label, i.e., Q is narrow. Similarly, the \textit{inverse KL} divergence is given by:
\[\beta\text{-}KL = D_{\text{KL}}(Q \parallel P) =  \sum_{x \in \mathcal{X}} Q(x) \log \left(\frac{Q(x)}{P(x)}\right)\]
$\beta$-KL is larger in the opposite scenario---when P is narrow, i.e., humans mostly agree on a label, but Q is wide, i.e., model's label probabilities are widely distributed, or Q has a peak on another label that humans do not agree. Therefore, the $\alpha$-KL is mainly influenced by instances with low human agreement but high model confidence. In contrast, the $\beta$-KL indicates situations where humans mostly agree, but the model lacks confidence. Both types of situations can reveal discrepancies in decision-making similarities, hence valuable for the user study.

\paragraph{Jensen-Shannon Divergence:} 
Although similar to KL-divergence, it is symmetric and finite, which makes it more intuitive~\cite{nie-etal-2020-learn}. 
\[D_{\text{JSD}}(P \parallel Q) = \frac{1}{2} \left(D_{\text{KL}}\left(P \parallel \frac{P+Q}{2}\right) \right.\]
\[\left. + D_{\text{KL}}\left(Q \parallel \frac{P+Q}{2}\right)\right)\]
We use Jensen-Shannon Distance calculated as $\sqrt{D_{\text{JSD}}}$ in our analysis. 
JSD considers the mixture distribution of P and Q and calculates forward KL divergences from both distributions. Consequently, one would expect JSD to reflect both $\alpha$ and $\beta$ KL. Hence, we anticipate JSD to strike a balance, effectively capturing the impact of instances with high human agreement but low machine confidence and vice versa. Therefore we choose JSD as the primary distance metric for the user study.

\section{User Study}
\label{sec:user_study}
We conduct a user study with 100 college-educated annotators with good command of English, recruited from Prolific~\footnote{\url{https://app.prolific.co/}}. We require annotators to have a minimum approval rate of 90\%. Further, we have set the college level as a filtering criterion. We have eliminated the participants from 159 to 100. The median time taken for the study was 35 minutes. Our average payment for each annotation was £4.00~\footnote{The minimum wage in the UK is £10.42/hr.}. Furthermore, we filtered out user results with an accuracy below 0.60 during the annotation phase.

\subsection{Subset selection} 
\label{ssec:subset}

To begin, we first filter instances where two or more models have identical labels and a confidence score exceeding 0.95. This results in the removal of 43 out of 90 instances. As a result, the accuracy of the LSTM and RoBERTa models falls short of aggregated human performance. To maintain the accuracy ranking, we introduce an additional 7 instances where the models succeed while the aggregated human answer fails. The accuracy variation between two sets can be found in Table \ref{table:accr-results}.

\begin{table}[!h]
    \begin{adjustbox}{}
    \begin{tabular}{|l|l|}
    \hline
    \textbf{Model}       & \textbf{Acc$_{original/subset}$} \\
    \hline
    Random Baseline      & 0.35 / 0.26                      \\
    TF-IDF               & 0.51 / 0.28                      \\
    LSTM (ESIM)          & 0.84 / 0.72                      \\
    RoBERTa              & 0.84 / 0.72                      \\
    Davinci              & 0.59 / 0.42                      \\
    \hline
    Human & 0.81 / 0.72                      \\
    \hline
    \end{tabular}
    \caption{\label{table:accr-results} Model accuracy variations with subset selection. Human: Aggregated human annotations.}
    \end{adjustbox}
\end{table}

\subsection{Design of the User Study}

\begin{figure}[!h]
    \includegraphics[scale=0.35]{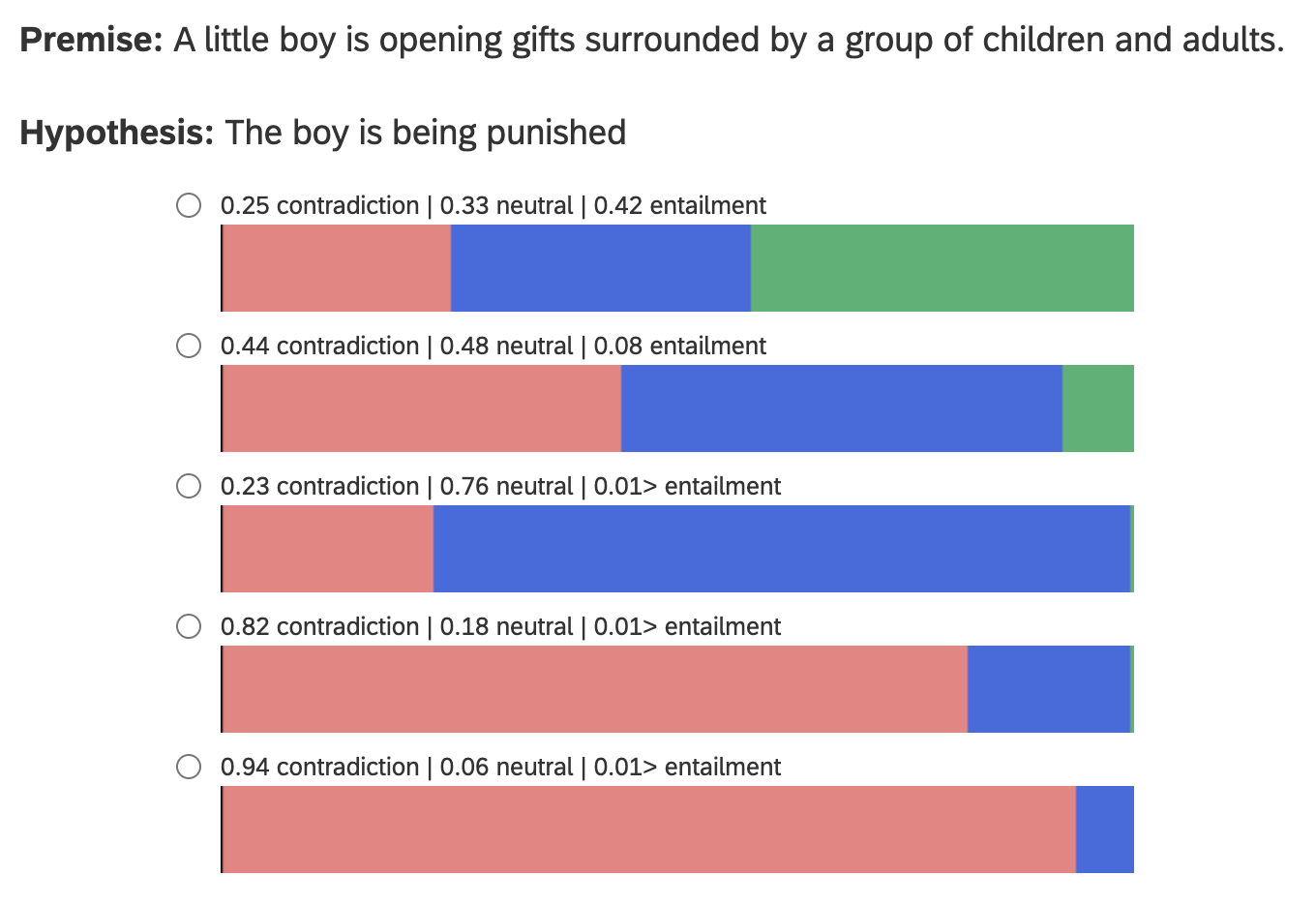}
    \caption{The annotation task is framed as a multiple-choice question answering problem, where the available options correspond to label predictions generated by the selected models.}
  \label{fig:userstudy-question}
\end{figure}

Our user study contains four stages: i) training, ii) quiz iii) annotation, and iv) dynamic survey. During \textbf{training phase}, users are provided with guidelines, which contain the general flow of the study with detailed instructions and a detailed explanation of the annotation task with examples. Full annotation guidelines can be found in Appendix A. In part ii), participants answer ten questions and then see the correct answers with explanations for training purposes. iii) During the \textbf{annotation phase}, the participants are shown the selected premise-hypothesis pairs, along with the label distributions from the selected models (see \S\ref{ssec:models})  as shown in Fig \ref{fig:userstudy-question}. The users are instructed to choose the \textit{closest agent} by means of the label distributions. The order of the models is shuffled each time to prevent users from finding shortcuts. Only one option can be chosen. Participants can skip the question and go back to the question. They are also asked to rate the difficulty of the question from 1 to 10.

After the annotation phase, we calculate the most similar model to the user with the JSD metric. This is because JSD contains both the $\alpha$ and the $\beta$-KL calculations, penalizing models that are overly confident on a single label w.r.t the users (high $\alpha$-KL), and vice versa (high $\beta$-KL). We calculate the distance between the user's answer and all the other label distributions. Then, we sum the distances for each model and identify the most similar one to the user, which we refer to as the \textit{aligned model}.

Then, in the \textbf{dynamic survey phase}, the users are asked to review their own selections side-by-side with their aligned model. They can investigate four categories: (A) agreements with the aligned model: the common successes and failures, 
and (B) the disagreements with the aligned model: the cases where the model succeeded and the user failed, and vice versa. 
The details are given in App.~\ref{sec:appendix}. After they revise the similarities and dissimilarities, they are asked several questions regarding the model behavior and the following outcomes:  (1) Collaboration: ``Would you collaborate with this agent to accomplish the task?'',  (2) Cognitive Trust: ``On a scale of 1 to 5, how much do you trust the agent to make rational decisions?'' For collaboration, the choices are provided in verbal likert scale format, including ``definitely not'', ``probably not'', ``neutral'', ``probably yes'', and ``definitely yes''. Finally, to validate our assumptions of the task, we ask several task-related questions, such as the difficulty and subjectivity of the task, and skills needed to solve the task, such as emotional awareness, logical thinking, and being a native English speaker. To date, we have not exploited this information, and we will do so in future work, especially when there are more than SNLI tasks to perform. A full list of the questions and related figures can be found in Appendix.

\section{Results}
\label{sec:results}
\begin{table*}[!htp]
\centering

\begin{tabular}{ll ll ll l}
\toprule
\multirow{2}{*}{\textbf{\#p}}  &
\multirow{2}{*}
{\textbf
    {\begin{tabular}[c]
        {@{}l@
        {}}
        Aligned\\ model
    \end{tabular}
    }
} &

\multirow{2}{*}
{\textbf
    {\begin{tabular}[c]
        {@{}l@{}}$\alpha$-KL
    \end{tabular}
    }
} &
\multirow{2}{*}
{\textbf{
    \begin{tabular}[c]
    {@{}l@{}}$\beta$-KL
    \end{tabular}
    }
}     &
\multirow{2}{*}
{\textbf{
    \begin{tabular}[c]
    {@{}l@{}} JSD\end{tabular}
    }
} &

\multirow{2}{*}
{\textbf{
    \begin{tabular}[c]
    {@{}l@{}}Collaboration\\  rating\end{tabular}
    }
} & 
\multirow{2}{*}
{\textbf{
    \begin{tabular}[c]
    {@{}l@{}}Cognitive trust\\  rating\end{tabular}
    }
}\\ \\
\midrule
60 & RoBERTa  & 0.53 & 1.21 & 0.23 & 3.9 & 3.36             \\
34 & LSTM     & 0.64 & 1.20 & 0.22 & 3.7 & \textbf{3.41}    \\
6  & Davinci  & 0.81 & 0.98 & 0.26 & \textbf{4.1} & 2.66    \\
\bottomrule
100 & average & 0.64 & 1.19 & 0.23 & 3.88 & 3.34            \\
\end{tabular}

\caption{\label{table:ranking-results} User study rating results over 5 points. \#p: Number of participants aligned with the model. Rows show the averaged scores for the model calculated over the users aligned with it. The final row shows the averaged scores over all models.}
\end{table*}

\begin{figure}[hb!]
    \includegraphics[width=90mm,scale=0.3]{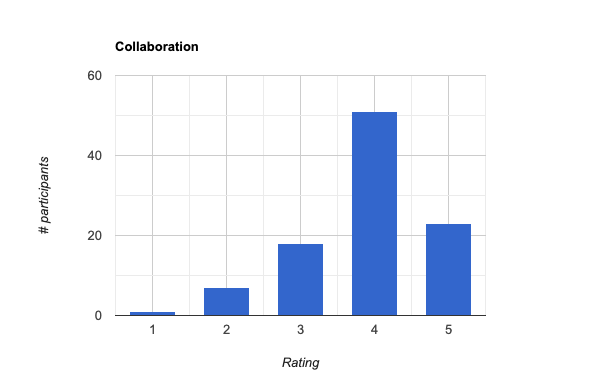}
    \caption{Collaboration ratings among users. Over half of the participants give 4/5 ratings for collaboration with their aligned model.}
  \label{fig:collab_histo.png}
\end{figure}

\begin{figure}[hb!]
    \includegraphics[width=90mm, scale=0.3]{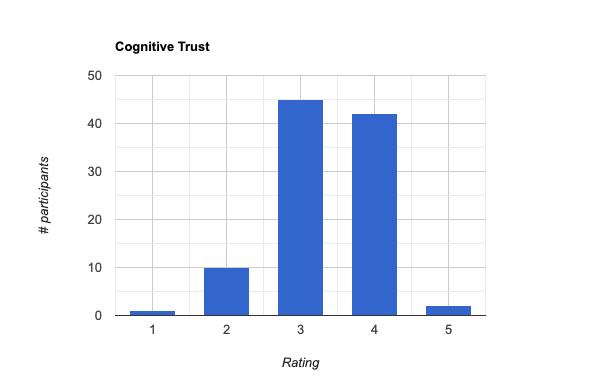}
    \caption{Cognitive Trust ratings among users. Participants are distributed across 3 and 4/5 ratings for their aligned model.}
  \label{fig:trust_histo.png}
\end{figure}

Our main results are given in Table~\ref{table:ranking-results}. Here, the collaboration and cognitive trust scores are averaged over the users that are \textit{aligned} with the model (shown with the rows). Then each score is again averaged to get a final score for each aspect. As given in Table \ref{table:ranking-results}, the majority of the users (60\%) are aligned with RoBERTa, followed by LSTM (34\%) and Davinci (6\%). None of the users are found similar to the random baseline or the tf-idf model, validating the suitability of the chosen metric and the quality of the user study. We find that the average collaboration score (3.88/5) is considerably higher than the average cognitive trust (3.34/5) score (the full distributions can be seen in Figures \ref{fig:collab_histo.png} and \ref{fig:trust_histo.png}). Finally, in Table \ref{table:hard-label-results}, we provide additional information on the models, users and their decision variations. We elaborate on the results below separately for collaboration and cognitive trust.

\paragraph{Collaboration}
According to Table~\ref{table:ranking-results}, Davinci has the highest collaboration score despite having the lowest cognitive trust score. It has also the lowest $\beta$-KL divergence to its aligned users. As discussed in \S\ref{sec:similarity}, $\beta$-KL gets higher when the user predicts a label with high confidence, but the model has less confidence. Therefore, Davinci most effectively fulfills confidence expectations of its aligned users for their answers. In contrast, users aligned with the RoBERTa and LSTM models exhibit higher $\beta$-KL values. That means users are more confident than their models in their predictions. They also have lower collaboration ratings, suggesting that agreeing on the same answer with high confidence is a critical DMS feature for collaboration. As a result, as seen in Table \ref{table:hard-label-results}, Davinci and its corresponding users establish the most confident aligned user-machine pairs. 

\paragraph{Cognitive Trust}
 
We expect participants to have a sense of the rational characteristics of the models after the user study. This is because the disagreement and agreement cases are shown exclusively to them during the dynamic evaluation phase. Our results in Table \ref{table:ranking-results} shows that LSTM has the highest cognitive trust scores, followed closely by RoBERTa. However, we notice a significant decrease in trust for Davinci. As given in Table~\ref{table:ranking-results}, Davinci suffers from high $\alpha$-KL. As discussed in \S\ref{sec:similarity}, $\alpha$-KL gets higher when a model is overconfident in its predicted label, but the user has low confidence in that label; instead, user confidence spreads over the labels. Davinci has also relatively lower accuracy than the average accuracy of its aligned users as shown in Table~\ref{table:hard-label-results}. This also indicates the overconfidence of the model in wrong labels for most of the questions. We think that observing this dissimilarity between themselves and models damages cognitive trust of users in the machine. We further observe that,  JSD, where combines the effects of both  $\alpha$-KL and 
$\beta$-KL divergences give the same model rankings (LSTM-RoBERTa-Davinci) for cognitive trust.  Therefore, it might be more advantageous to incorporate both divergences and solely rely on JSD for assessing cognitive trust. However more comprehensive model comparisons are needed to draw a concrete solution, which is not feasible in this user study.

\ggrev{\paragraph{Summary} To summarize, our findings demonstrate that for collaboration purposes, participants expect a specific type of similarity with models: to have high confidence agreement on their answers, \textbf{as indicated by $\beta$-KL}.
Task delegation has links with collaboration and cognitive trust such as high confidence agreement; however, it additionally requires \textbf{high accuracy}. Hence, combining \textbf{$\beta$-KL with accuracy} could be valuable for measuring and improving delegation. Regarding cognitive trust, our results indicate that having high similarity in confidence levels for a given label is important, similar to collaboration and delegation as indicated by \textbf{lower $\beta$-KL}. However, it is also crucial to show similarity in low confidence and avoid being overconfident, as indicated by \textbf{lower $\alpha$-KL}. In addition, model accuracy also plays a significant role in cognitive trust. Therefore, we suggest that both \textbf{JSD and accuracy} are considered to measure/improve it. For perceived human-likeness, we recommend that \textbf{JSD} is used as it takes both similarity directions into account.}{}
To summarize, we observe that participants tend to collaborate with their most similar models; and they give higher collaboration ratings when they share high confidence in answers with their models, as indicated by \textbf{lower $\beta$-KL}. However, this is not the case for cognitive trust. Participants give higher trust ratings when they share low confidence in their answers and avoid being overconfident, as indicated by \textbf{lower $\alpha$-KL} and \textbf{JSD}. The results also demonstrate that individuals may assign higher collaboration ratings to models, even if they have low cognitive trust to the model.

\begin{table*}[h!]
\begin{adjustbox}{width=\textwidth}
\begin{tabular}{lllllllll}
\hline
\multirow{2}{*}{\textbf{\begin{tabular}[c]{@{}l@{}}Aligned\\ model\end{tabular}}} &
\multirow{2}{*}{\textbf{\begin{tabular}[c]{@{}l@{}}Model\\ acc.\end{tabular}}} &
\multirow{2}{*}{\textbf{\begin{tabular}[c]{@{}l@{}}User\\ acc.\end{tabular}}} &
\multirow{2}{*}{\textbf{\begin{tabular}[c]{@{}l@{}}Model\\ conf.\end{tabular}}} &
\multirow{2}{*}{\textbf{\begin{tabular}[c]{@{}l@{}}User\\ conf.\end{tabular}}} &
\multirow{2}{*}{\textbf{\begin{tabular}[c]{@{}l@{}}Soft label\\ agreement\end{tabular}}} &
\multirow{2}{*}{\textbf{\begin{tabular}[c]{@{}l@{}}Hard label\\ agreement\end{tabular}}} &
\multirow{2}{*}{\textbf{\begin{tabular}[c]{@{}l@{}}Selected\\ higher conf.\end{tabular}}} &
\multirow{2}{*}{\textbf{\begin{tabular}[c]{@{}l@{}}Selected\\ lower conf.\end{tabular}}} \\
&&&&&&&&\\
\hline
RoBERTa & 0.72 & 0.69 & 0.80 & 0.86 & 0.25 & \textbf{0.71} & 0.40 & 0.05 \\
LSTM & 0.72 & 0.70 & 0.83 & 0.84 & 0.39 & 0.69 & 0.16 & 0.14 \\
Davinci & 0.42 & 0.66 & \textbf{0.94} & \textbf{0.91} & \textbf{0.53} & 0.64 & 0.05 & 0.05 \\
\hline
\end{tabular}
\end{adjustbox}
\caption{\label{table:hard-label-results} User-model performances and selection details. Soft label agreement: Number of times user-model agree on soft label. Hard label agreement: Number of times user-model agree on hard label. Selected higher/lower conf: Number of times users selecting different model with higher/lower confidence on the same label. }
\end{table*}

\section{Discussion and Analysis}
\label{sec:discussion}
\noindent\textbf{(1) When multiple models share the same final labels, which one do people typically choose?} We analyze the test instances where multiple models provide the same hard labels but distinct label distributions. 
We then compare how frequently the users choose the highly confident model versus the less confident model. The majority of participants (67\%) choose the most confident model answer for 97\% of the cases. This result supports the findings of \citet{vodrahalli2022uncalibrated}, which suggests individuals accept solutions from highly confident models. We also see that as average user accuracy decreases, users tend to choose alternative models with the same label but lower confidence. So their answers are distributed more evenly across other labels. To quantify, we calculate Pearson's r between the number of times users selected the models and the average user accuracy on that question. With \textit{r} -0.41, the number of times selecting low confident models increases as average user accuracy decreases. This result suggests that humans will often choose the most confident model, but the label distribution of a model will also be critical in cases when they are uncertain. 

\noindent\textbf{(2) When do people not choose their most similar model?} 
We first calculate the number of times the user directly chooses its matched model answer.
In Table~\ref{table:hard-label-results}, we see that the majority of Davinci users have comparable confidence levels and agree with their model on 53\% of the cases. Therefore only for 10\% of the questions, users prefer other models' answer with the same label. However, overall they have the lowest agreement with their model. For RoBERTa, aligned users choose higher confident model answers in 40\% of the cases. 
RoBERTa has the lowest average confidence and its users tend to select more confident models. As for LSTM, the agreement between users and the model is nearly the same as with RoBERTa. Nevertheless, in most cases, people choose the LSTM model itself, but in 14\% of the cases, they choose models with lower confidence.

\paragraph{(3) What is the relation between collaboration and cognitive trust?}
We calculate Pearson's \textit{r} between collaboration and cognitive trust ratings of users. We first omit two outliers from the data. Then, over 98 user data ratings, with r=0.3, p=0.002, we see a weak positive correlation between them. This result once again suggests that cognitive trust may not be a necessity for high collaboration rates. 

\section{Related Work}
\label{sec:rel_work}
\paragraph{Factors in human-AI collaboration:} 
The factors to consider in HAI can change based on the type of AI-enabled systems, such as whether it is a physical machine, such as a robot, or a virtual agent like today's popular NLP models ~\cite{lee2018understanding,gillath2021attachment}. \citet{araujo2020ai} study the human factor by analyzing different user groups in automated decision-making with AI. They consider privacy concerns, age, decision-making type, AI role, and other features and investigate AI's perceived justice, usefulness, and risk. \citet{bao2021investigating, lee2018understanding, gillath2021attachment} study the relationship between trust and human-AI collaboration. \citet{glikson2020human} reviewed the trust factors and identified two dimensions of trust: i) the \textit{cognitive trust} based on the ``good rationality skills'' of the object of trust, and ii) the \textit{emotional trust} based on the positive effect for the object of trust~\cite{lewis1985trust}. 
Another line of work studies AI-advised decision-making scenarios.
For instance, \citet{bansal2019beyond} find that accuracy is not the primary metric for achieving the best human human-machine team performance. \citet{bansal2021most} further optimize AI models 
with human preferences and suggests that models should minimize the number of solutions that provide low-confidence accurate answers in favor of high-confidence accurate answers for better collaboration. \citet{vodrahalli2022uncalibrated} report similar findings and show that high confidence in AI predictions increases human intention to incorporate them in their decision-making. 
Moreover, \citet{bansal2019beyond} demonstrate that human awareness of AI errors enhances overall human-AI team accuracy. Thus, establishing clear and predictable AI error boundaries is crucial for improving collaboration with AI.

\paragraph{Soft-labels:} 
\citet{pavlick-kwiatkowski-2019-inherent} demonstrate that the variation of human annotations in the SNLI dataset might contain valuable signals and should not be treated as noise. It is a pattern observed when the number of annotators increases and can be identified within Gaussian mixture models. \citet{plank-2022-problem,uma2021learning} report the annotation variation and argue that this variation should not be discarded. Furthermore, this variation has been shown beneficial for optimizing models by using \textit{soft labels}---probability distributions over human annotations---to achieve better performance with multi-task learning setups instead of single \textit{hard labels}~\cite{fornaciari-etal-2021-beyond,davani-etal-2022-dealing, Meissner2021EmbracingAS}. 

\paragraph{Sample size in user studies:}
In Human-Computer Interaction literature on human factors, a sample size of 100 is commonly applicable by local standards \cite{10.1145/2858036.2858498}, where many studies involve even fewer than 100 participants. For instance, \citet{10.1145/3544548.3580794} surveyed 111 participants to explore the impact of AI knowledge on enhancing AI delegation. Similarly, \citet{10.1145/3544548.3580905} examined the efficacy of AI augmentation for real-time mathematical tasks, assessing user anxiety through a Likert scale with a cohort of 80 participants. \citet{10.1145/3544548.3581529} involved 81 participants to study AI's acceptance in human decision-makings. \citet{10.1145/3544548.3580917} investigates the physiological effects of cognitive biases with 33 participants. In the context of designing an editing tool for conceptual diagrams, \citet{10.1145/3544548.3580676} engaged 44 participants, while \citet{10.1145/3544548.3581282} study involved 78 individuals to compare AI recommendations in human-AI teamwork. Furthermore, the ChaosNLI dataset~\cite{nie-etal-2020-learn} collected opinions from 100 participants to handle the diversity in human reasoning. Finally, \citet{Conroy2016TheRS} show that a sample size of 100 can provide a ±10\% margin of uncertainty when studying group characteristics or behavior.

\section{Conclusion and Future Work}

In this work, we design a novel experimental setup on a textual entailment task to analyze the decision-making similarities between AI agents and humans, and their impact on collaboration and cognitive trust. To do so, we train a set of diverse models ranging from tf-idf to RoBERTa with varying accuracies and confidence levels. We then identify and calculate a set of divergence scores to measure the distance between human-AI label distributions. Finally, we design and conduct a comprehensive four-staged user study where the users are tasked to choose a \textit{label distribution} instead of a single label and are matched/aligned with an agent using the identified metrics. We survey the users about their aligned model (e.g., whether they would collaborate or trust the model) and find that different divergence metrics on soft-labels, i.e., $\alpha$-KL, $\beta$-KL, and JSD) contribute differently to collaboration preference and cognitive trust.

Our results show that people tend to collaborate with models when their decision-making processes are similar. However, they might not fully trust these models at the same level. Nevertheless, people are still likely to collaborate with models even if they show low cognitive trust. A specific type of DMS between user and model affects collaboration likelihood: agreeing on the same answer (right or wrong) with high confidence. This type of DMS seems necessary but insufficient for cognitive trust likelihood. A trust may further require another type of DMS,  i.e., the model avoids being overconfident in cases where users have low confidence.

Future work will focus on exploiting the user study results to optimize the models e.g., via fine-tuning, to improve specifically collaboration or cognitive trust. Furthermore, the proposed divergence scores along with the user study can be deployed to evaluate the models on those aspects before deployment.

\section*{Limitations}

Our design choice in user study led the models to receive feedback from different number of users. The study may be increased to a higher number of participants to have more robust results. Regarding datasets that could be used in model optimization or evaluation, we examined the commonly used method in literature, which involves aggregating diverse human annotations to capture collective human opinions (see Appendix B). However, aggregated group annotations do not always reflect the user's real divergence from the models. Therefore, we also raise the question regarding the reliance on aggregated human annotations as ground truth to reflect accurate human opinions on model interactions. Future research could also focus on that and data collection to accurately reflect human opinions for models at interaction.

\section*{Ethics Statement}

In this study, with the aim of understanding the relationship between evaluation metrics and human perceptions of NLP models after an interaction on an NLI task, we studied 100 people through an online platform. Though we gained demographic data like age, education, gender, and being a native English speaker, we did not collect any information that could reveal users' identities. We made sure to keep everything anonymous so that we can share results publicly. While we performed an analysis of human behavior, we did not make any specific positive or negative claims about the participants or people in general.

\section*{Acknowledgements}
The first author has been supported by KUIS AI Center fellowship. This work has been supported by the Scientific and Technological Research Council of Türkiye~(TÜBİTAK) as part of the project ``Automatic Learning of Procedural Language from Natural Language Instructions for Intelligent Assistance'' with the number 121C132. We also gratefully acknowledge KUIS AI Center for providing computational support. We thank our reviewers and the members of GGLab who helped us improve this paper.

\bibliography{custom}

\begin{thebibliography}{46}
\expandafter\ifx\csname natexlab\endcsname\relax\def\natexlab#1{#1}\fi

\bibitem[{Abdelghani et~al.(2023)Abdelghani, Wang, Yuan, Wang, Lucas, Sauz{\'e}on, and Oudeyer}]{abdelghani2023gpt}
Rania Abdelghani, Yen-Hsiang Wang, Xingdi Yuan, Tong Wang, Pauline Lucas, H{\'e}l{\`e}ne Sauz{\'e}on, and Pierre-Yves Oudeyer. 2023.
\newblock Gpt-3-driven pedagogical agents to train children’s curious question-asking skills.
\newblock \emph{International Journal of Artificial Intelligence in Education}, pages 1--36.

\bibitem[{Araujo et~al.(2020)Araujo, Helberger, Kruikemeier, and De~Vreese}]{araujo2020ai}
Theo Araujo, Natali Helberger, Sanne Kruikemeier, and Claes~H De~Vreese. 2020.
\newblock In ai we trust? perceptions about automated decision-making by artificial intelligence.
\newblock \emph{AI \& society}, 35:611--623.

\bibitem[{Bansal et~al.(2021{\natexlab{a}})Bansal, Nushi, Kamar, Horvitz, and Weld}]{bansal2021most}
Gagan Bansal, Besmira Nushi, Ece Kamar, Eric Horvitz, and Daniel~S Weld. 2021{\natexlab{a}}.
\newblock Is the most accurate ai the best teammate? optimizing ai for teamwork.
\newblock In \emph{Proceedings of the AAAI Conference on Artificial Intelligence}, volume~35, pages 11405--11414.

\bibitem[{Bansal et~al.(2019)Bansal, Nushi, Kamar, Lasecki, Weld, and Horvitz}]{bansal2019beyond}
Gagan Bansal, Besmira Nushi, Ece Kamar, Walter~S Lasecki, Daniel~S Weld, and Eric Horvitz. 2019.
\newblock Beyond accuracy: The role of mental models in human-ai team performance.
\newblock In \emph{Proceedings of the AAAI conference on human computation and crowdsourcing}, volume~7, pages 2--11.

\bibitem[{Bansal et~al.(2021{\natexlab{b}})Bansal, Wu, Zhou, Fok, Nushi, Kamar, Ribeiro, and Weld}]{bansal2021does}
Gagan Bansal, Tongshuang Wu, Joyce Zhou, Raymond Fok, Besmira Nushi, Ece Kamar, Marco~Tulio Ribeiro, and Daniel~S. Weld. 2021{\natexlab{b}}.
\newblock \href {http://arxiv.org/abs/2006.14779} {Does the whole exceed its parts? the effect of ai explanations on complementary team performance}.

\bibitem[{Bao et~al.(2021)Bao, Cheng, De~Vreede, and De~Vreede}]{bao2021investigating}
Ying Bao, Xusen Cheng, Triparna De~Vreede, and Gert-Jan De~Vreede. 2021.
\newblock Investigating the relationship between ai and trust in human-ai collaboration.

\bibitem[{Barke et~al.(2023)Barke, James, and Polikarpova}]{barke2023grounded}
Shraddha Barke, Michael~B James, and Nadia Polikarpova. 2023.
\newblock Grounded copilot: How programmers interact with code-generating models.
\newblock \emph{Proceedings of the ACM on Programming Languages}, 7(OOPSLA1):85--111.

\bibitem[{Boonprakong et~al.(2023)Boonprakong, Chen, Davey, Tag, and Dingler}]{10.1145/3544548.3580917}
Nattapat Boonprakong, Xiuge Chen, Catherine Davey, Benjamin Tag, and Tilman Dingler. 2023.
\newblock \href {https://doi.org/10.1145/3544548.3580917} {Bias-aware systems: Exploring indicators for the occurrences of cognitive biases when facing different opinions}.
\newblock In \emph{Proceedings of the 2023 CHI Conference on Human Factors in Computing Systems}, CHI '23, New York, NY, USA. Association for Computing Machinery.

\bibitem[{Bowman et~al.(2015)Bowman, Angeli, Potts, and Manning}]{bowman-etal-2015-large}
Samuel~R. Bowman, Gabor Angeli, Christopher Potts, and Christopher~D. Manning. 2015.
\newblock \href {https://doi.org/10.18653/v1/D15-1075} {A large annotated corpus for learning natural language inference}.
\newblock In \emph{Proceedings of the 2015 Conference on Empirical Methods in Natural Language Processing}, pages 632--642, Lisbon, Portugal. Association for Computational Linguistics.

\bibitem[{Brown et~al.(2020)Brown, Mann, Ryder, Subbiah, Kaplan, Dhariwal, Neelakantan, Shyam, Sastry, Askell, Agarwal, Herbert-Voss, Krueger, Henighan, Child, Ramesh, Ziegler, Wu, Winter, Hesse, Chen, Sigler, Litwin, Gray, Chess, Clark, Berner, McCandlish, Radford, Sutskever, and Amodei}]{brown2020language}
Tom~B. Brown, Benjamin Mann, Nick Ryder, Melanie Subbiah, Jared Kaplan, Prafulla Dhariwal, Arvind Neelakantan, Pranav Shyam, Girish Sastry, Amanda Askell, Sandhini Agarwal, Ariel Herbert-Voss, Gretchen Krueger, Tom Henighan, Rewon Child, Aditya Ramesh, Daniel~M. Ziegler, Jeffrey Wu, Clemens Winter, Christopher Hesse, Mark Chen, Eric Sigler, Mateusz Litwin, Scott Gray, Benjamin Chess, Jack Clark, Christopher Berner, Sam McCandlish, Alec Radford, Ilya Sutskever, and Dario Amodei. 2020.
\newblock \href {http://arxiv.org/abs/2005.14165} {Language models are few-shot learners}.

\bibitem[{Caine(2016)}]{10.1145/2858036.2858498}
Kelly Caine. 2016.
\newblock \href {https://doi.org/10.1145/2858036.2858498} {Local standards for sample size at chi}.
\newblock In \emph{Proceedings of the 2016 CHI Conference on Human Factors in Computing Systems}, CHI '16, page 981–992, New York, NY, USA. Association for Computing Machinery.

\bibitem[{Chen et~al.(2016)Chen, Zhu, Ling, Wei, Jiang, and Inkpen}]{chen2016enhanced}
Qian Chen, Xiaodan Zhu, Zhenhua Ling, Si~Wei, Hui Jiang, and Diana Inkpen. 2016.
\newblock Enhanced lstm for natural language inference.
\newblock \emph{arXiv preprint arXiv:1609.06038}.

\bibitem[{Conroy(2016)}]{Conroy2016TheRS}
Ron{\'a}n~M. Conroy. 2016.
\newblock \href {https://api.semanticscholar.org/CorpusID:173172014} {The rcsi sample size handbook}.

\bibitem[{Davani et~al.(2022)Davani, D{\'\i}az, and Prabhakaran}]{davani-etal-2022-dealing}
Aida~Mostafazadeh Davani, Mark D{\'\i}az, and Vinodkumar Prabhakaran. 2022.
\newblock \href {https://doi.org/10.1162/tacl_a_00449} {Dealing with disagreements: Looking beyond the majority vote in subjective annotations}.
\newblock \emph{Transactions of the Association for Computational Linguistics}, 10:92--110.

\bibitem[{Dijkstra et~al.(2022)Dijkstra, Gen{\c{c}}, Kayal, Kamps et~al.}]{dijkstra2022reading}
Ramon Dijkstra, Z{\"u}lk{\"u}f Gen{\c{c}}, Subhradeep Kayal, Jaap Kamps, et~al. 2022.
\newblock Reading comprehension quiz generation using generative pre-trained transformers.

\bibitem[{Fornaciari et~al.(2021)Fornaciari, Uma, Paun, Plank, Hovy, and Poesio}]{fornaciari-etal-2021-beyond}
Tommaso Fornaciari, Alexandra Uma, Silviu Paun, Barbara Plank, Dirk Hovy, and Massimo Poesio. 2021.
\newblock \href {https://doi.org/10.18653/v1/2021.naacl-main.204} {Beyond black {\&} white: Leveraging annotator disagreement via soft-label multi-task learning}.
\newblock In \emph{Proceedings of the 2021 Conference of the North American Chapter of the Association for Computational Linguistics: Human Language Technologies}, pages 2591--2597, Online. Association for Computational Linguistics.

\bibitem[{Gillath et~al.(2021)Gillath, Ai, Branicky, Keshmiri, Davison, and Spaulding}]{gillath2021attachment}
Omri Gillath, Ting Ai, Michael~S Branicky, Shawn Keshmiri, Robert~B Davison, and Ryan Spaulding. 2021.
\newblock Attachment and trust in artificial intelligence.
\newblock \emph{Computers in Human Behavior}, 115:106607.

\bibitem[{Glikson and Woolley(2020)}]{glikson2020human}
Ella Glikson and Anita~Williams Woolley. 2020.
\newblock Human trust in artificial intelligence: Review of empirical research.
\newblock \emph{Academy of Management Annals}, 14(2):627--660.

\bibitem[{Howard et~al.(2023)Howard, Hope, and Gerada}]{howard2023chatgpt}
Alex Howard, William Hope, and Alessandro Gerada. 2023.
\newblock Chatgpt and antimicrobial advice: the end of the consulting infection doctor?
\newblock \emph{The Lancet Infectious Diseases}, 23(4):405--406.

\bibitem[{Jeblick et~al.(2023)Jeblick, Schachtner, Dexl, Mittermeier, St{\"u}ber, Topalis, Weber, Wesp, Sabel, Ricke et~al.}]{jeblick2023chatgpt}
Katharina Jeblick, Balthasar Schachtner, Jakob Dexl, Andreas Mittermeier, Anna~Theresa St{\"u}ber, Johanna Topalis, Tobias Weber, Philipp Wesp, Bastian~Oliver Sabel, Jens Ricke, et~al. 2023.
\newblock Chatgpt makes medicine easy to swallow: an exploratory case study on simplified radiology reports.
\newblock \emph{European Radiology}, pages 1--9.

\bibitem[{Kocielnik et~al.(2019)Kocielnik, Amershi, and Bennett}]{kocielnik2019will}
Rafal Kocielnik, Saleema Amershi, and Paul~N Bennett. 2019.
\newblock Will you accept an imperfect ai? exploring designs for adjusting end-user expectations of ai systems.
\newblock In \emph{Proceedings of the 2019 CHI Conference on Human Factors in Computing Systems}, pages 1--14.

\bibitem[{Kullback and Leibler(1951)}]{kullback1951information}
Solomon Kullback and Richard~A Leibler. 1951.
\newblock On information and sufficiency.
\newblock \emph{The annals of mathematical statistics}, 22(1):79--86.

\bibitem[{Lee(2018)}]{lee2018understanding}
Min~Kyung Lee. 2018.
\newblock Understanding perception of algorithmic decisions: Fairness, trust, and emotion in response to algorithmic management.
\newblock \emph{Big Data \& Society}, 5(1):2053951718756684.

\bibitem[{Lewis and Weigert(1985)}]{lewis1985trust}
J~David Lewis and Andrew Weigert. 1985.
\newblock Trust as a social reality.
\newblock \emph{Social forces}, 63(4):967--985.

\bibitem[{Lin(1991)}]{Lin1991DivergenceMB}
Jianhua Lin. 1991.
\newblock Divergence measures based on the shannon entropy.
\newblock \emph{IEEE Trans. Inf. Theory}, 37:145--151.

\bibitem[{Liu et~al.(2023)Liu, Wang, and Liu}]{liu2023utility}
Jialin Liu, Changyu Wang, and Siru Liu. 2023.
\newblock Utility of chatgpt in clinical practice.
\newblock \emph{Journal of Medical Internet Research}, 25:e48568.

\bibitem[{Liu et~al.(2019)Liu, Ott, Goyal, Du, Joshi, Chen, Levy, Lewis, Zettlemoyer, and Stoyanov}]{Liu2019RoBERTaAR}
Yinhan Liu, Myle Ott, Naman Goyal, Jingfei Du, Mandar Joshi, Danqi Chen, Omer Levy, Mike Lewis, Luke Zettlemoyer, and Veselin Stoyanov. 2019.
\newblock Roberta: A robustly optimized bert pretraining approach.
\newblock \emph{ArXiv}, abs/1907.11692.

\bibitem[{Meissner et~al.(2021)Meissner, Thumwanit, Sugawara, and Aizawa}]{Meissner2021EmbracingAS}
Johannes~Mario Meissner, Napat Thumwanit, Saku Sugawara, and Akiko Aizawa. 2021.
\newblock Embracing ambiguity: Shifting the training target of nli models.
\newblock In \emph{Annual Meeting of the Association for Computational Linguistics}.

\bibitem[{Moore et~al.(2022)Moore, Nguyen, Bier, Domadia, and Stamper}]{moore2022assessing}
Steven Moore, Huy~A Nguyen, Norman Bier, Tanvi Domadia, and John Stamper. 2022.
\newblock Assessing the quality of student-generated short answer questions using gpt-3.
\newblock In \emph{European conference on technology enhanced learning}, pages 243--257. Springer.

\bibitem[{Nie et~al.(2020)Nie, Zhou, and Bansal}]{nie-etal-2020-learn}
Yixin Nie, Xiang Zhou, and Mohit Bansal. 2020.
\newblock \href {https://doi.org/10.18653/v1/2020.emnlp-main.734} {What can we learn from collective human opinions on natural language inference data?}
\newblock In \emph{Proceedings of the 2020 Conference on Empirical Methods in Natural Language Processing (EMNLP)}, pages 9131--9143, Online. Association for Computational Linguistics.

\bibitem[{OpenAI(2022)}]{openai_2022}
OpenAI. 2022.
\newblock \href {https://openai.com/blog/chatgpt} {Introducing chatgpt}.

\bibitem[{Pan et~al.(2023)Pan, Yu, He, and Shi}]{10.1145/3544548.3580676}
Lihang Pan, Chun Yu, Zhe He, and Yuanchun Shi. 2023.
\newblock \href {https://doi.org/10.1145/3544548.3580676} {A human-computer collaborative editing tool for conceptual diagrams}.
\newblock In \emph{Proceedings of the 2023 CHI Conference on Human Factors in Computing Systems}, CHI '23, New York, NY, USA. Association for Computing Machinery.

\bibitem[{Pavlick and Kwiatkowski(2019)}]{pavlick-kwiatkowski-2019-inherent}
Ellie Pavlick and Tom Kwiatkowski. 2019.
\newblock \href {https://doi.org/10.1162/tacl_a_00293} {Inherent disagreements in human textual inferences}.
\newblock \emph{Transactions of the Association for Computational Linguistics}, 7:677--694.

\bibitem[{Pinski et~al.(2023)Pinski, Adam, and Benlian}]{10.1145/3544548.3580794}
Marc Pinski, Martin Adam, and Alexander Benlian. 2023.
\newblock \href {https://doi.org/10.1145/3544548.3580794} {Ai knowledge: Improving ai delegation through human enablement}.
\newblock In \emph{Proceedings of the 2023 CHI Conference on Human Factors in Computing Systems}, CHI '23, New York, NY, USA. Association for Computing Machinery.

\bibitem[{Plank(2022)}]{plank-2022-problem}
Barbara Plank. 2022.
\newblock \href {https://aclanthology.org/2022.emnlp-main.731} {The {``}problem{''} of human label variation: On ground truth in data, modeling and evaluation}.
\newblock In \emph{Proceedings of the 2022 Conference on Empirical Methods in Natural Language Processing}, pages 10671--10682, Abu Dhabi, United Arab Emirates. Association for Computational Linguistics.

\bibitem[{Shin(2021)}]{SHIN2021102551}
Donghee Shin. 2021.
\newblock \href {https://doi.org/https://doi.org/10.1016/j.ijhcs.2020.102551} {The effects of explainability and causability on perception, trust, and acceptance: Implications for explainable ai}.
\newblock \emph{International Journal of Human-Computer Studies}, 146:102551.

\bibitem[{Skjuve()}]{skjuve4376834people}
Marita Skjuve.
\newblock Why people use chatgpt.
\newblock \emph{Available at SSRN 4376834}.

\bibitem[{Tretter et~al.(2023)Tretter, Platz, and Diefenbach}]{10.1145/3544548.3581529}
Stefan Tretter, Axel Platz, and Sarah Diefenbach. 2023.
\newblock \href {https://doi.org/10.1145/3544548.3581529} {Matching mind and method: Augmented decision-making with digital companions based on regulatory mode theory}.
\newblock In \emph{Proceedings of the 2023 CHI Conference on Human Factors in Computing Systems}, CHI '23, New York, NY, USA. Association for Computing Machinery.

\bibitem[{Uma et~al.(2021)Uma, Fornaciari, Hovy, Paun, Plank, and Poesio}]{uma2021learning}
Alexandra~N Uma, Tommaso Fornaciari, Dirk Hovy, Silviu Paun, Barbara Plank, and Massimo Poesio. 2021.
\newblock Learning from disagreement: A survey.
\newblock \emph{Journal of Artificial Intelligence Research}, 72:1385--1470.

\bibitem[{Vodrahalli et~al.(2022)Vodrahalli, Gerstenberg, and Zou}]{vodrahalli2022uncalibrated}
Kailas Vodrahalli, Tobias Gerstenberg, and James~Y Zou. 2022.
\newblock Uncalibrated models can improve human-ai collaboration.
\newblock \emph{Advances in Neural Information Processing Systems}, 35:4004--4016.

\bibitem[{Wang et~al.(2019)Wang, Pruksachatkun, Nangia, Singh, Michael, Hill, Levy, and Bowman}]{WangPNSMHLB19}
Alex Wang, Yada Pruksachatkun, Nikita Nangia, Amanpreet Singh, Julian Michael, Felix Hill, Omer Levy, and Samuel~R. Bowman. 2019.
\newblock \href {https://proceedings.neurips.cc/paper/2019/hash/4496bf24afe7fab6f046bf4923da8de6-Abstract.html} {Superglue: {A} stickier benchmark for general-purpose language understanding systems}.
\newblock In \emph{Advances in Neural Information Processing Systems 32: Annual Conference on Neural Information Processing Systems 2019, NeurIPS 2019, December 8-14, 2019, Vancouver, BC, Canada}, pages 3261--3275.

\bibitem[{Webson and Pavlick(2022)}]{webson2022promptbased}
Albert Webson and Ellie Pavlick. 2022.
\newblock \href {http://arxiv.org/abs/2109.01247} {Do prompt-based models really understand the meaning of their prompts?}

\bibitem[{Xu et~al.(2023)Xu, Lien, and H\"{o}llerer}]{10.1145/3544548.3581282}
Chengyuan Xu, Kuo-Chin Lien, and Tobias H\"{o}llerer. 2023.
\newblock \href {https://doi.org/10.1145/3544548.3581282} {Comparing zealous and restrained ai recommendations in a real-world human-ai collaboration task}.
\newblock In \emph{Proceedings of the 2023 CHI Conference on Human Factors in Computing Systems}, CHI '23, New York, NY, USA. Association for Computing Machinery.

\bibitem[{Xu and Zhang(2023)}]{10.1145/3544548.3580905}
Songlin Xu and Xinyu Zhang. 2023.
\newblock \href {https://doi.org/10.1145/3544548.3580905} {Augmenting human cognition with an ai-mediated intelligent visual feedback}.
\newblock In \emph{Proceedings of the 2023 CHI Conference on Human Factors in Computing Systems}, CHI '23, New York, NY, USA. Association for Computing Machinery.

\bibitem[{Yin et~al.(2019)Yin, Wortman~Vaughan, and Wallach}]{yin2019understanding}
Ming Yin, Jennifer Wortman~Vaughan, and Hanna Wallach. 2019.
\newblock Understanding the effect of accuracy on trust in machine learning models.
\newblock In \emph{Proceedings of the 2019 chi conference on human factors in computing systems}, pages 1--12.

\bibitem[{Zhang et~al.(2020)Zhang, Liao, and Bellamy}]{zhang2020effect}
Yunfeng Zhang, Q~Vera Liao, and Rachel~KE Bellamy. 2020.
\newblock Effect of confidence and explanation on accuracy and trust calibration in ai-assisted decision making.
\newblock In \emph{Proceedings of the 2020 conference on fairness, accountability, and transparency}, pages 295--305.

\end{thebibliography}
\bibliographystyle{acl_natbib}

\clearpage
\appendix
\clearpage
\label{sec:appendix}
\section{User Study}
\subsection{Guidelines}
\begin{minipage}{2\columnwidth}
\centering
\begin{tabular}{|p{0.9\columnwidth}|}
\hline
\textbf{Guidelines} \\
\hline
\textit{\textbf{Page 1.}} \\
In this survey, you will be shown a pair of sentences in which you should determine the inference relation between them.  The first sentence is a premise, and the second is a candidate hypothesis. There are three possible inference relations between sentences: entailment, neutral, or contradiction.\\
In the first part of the survey, you will be asked 10 test questions to confirm the task is understood. In the second part, you will be asked 50 questions from the task. In the final part, you will be asked task-related questions. The survey will take approximately 30 min. \\
Thank you for your participation.\\

\hline

\textit{\textbf{Page 2.}} \\
\textbf{Task Examples}: \\
 \\
\textbf{Entailment}: Premise sentence definitely implies the hypothesis.\\
e.g.:\\
Premise: “Two dogs are running through a field."\\
Hypothesis: “There are animals outdoors.”\\
Correct label: Entailment\\
\\
\textbf{Neutral}: Premise sentence might imply the hypothesis.\\
e.g.:\\
Premise: “Two dogs are running through a field."\\
Hypothesis: “Some puppies are running to catch a stick”\\
Correct label: Neutral\\
\\
\textbf{Contradiction}: Premise sentence definitely does not imply the hypothesis.\\
e.g.:\\
Premise: “Two dogs are running through a field."\\
Hypothesis: “The pets are sitting on a couch.”\\
Correct label: Contradiction. This is different from the \textit{maybe correct (neutral)} category because it’s impossible for the dogs to be both running and sitting.\\
\hline

\textit{\textbf{Page 3.}} \\
In the next part of the survey, you will answer 50 questions with multiple choices that include confidence-relation pairs. You can select the choice which reflects your confidence in labels. \\
\\
Premise:             <premise\_sentence>\\
Hypothesis:          <hypothesis\_sentence>\\
Choice format:       x\% Contradiction    | y\% Neutral   | z\% Entailment\\ 
e.g:                 0.55 Contradiction  | 0.35 Neutral | 0.10 Entailment \\

\hline
\end{tabular}
\captionof{table}{User study guideline}
\label{table:userstudy_guideline}
\end{minipage}

\clearpage
\subsection{Results Page}
\begin{minipage}{2\columnwidth}
\centering
\includegraphics[width=0.7\textwidth]{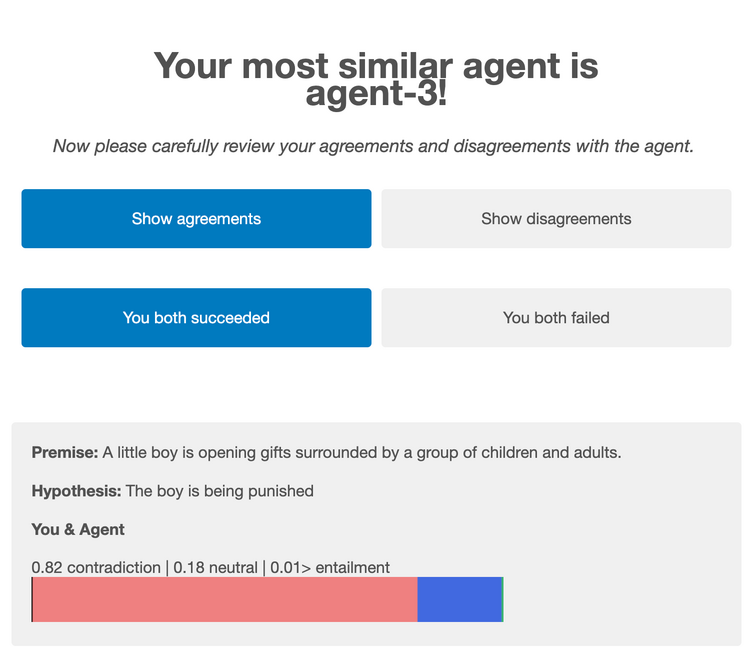}\\
(a) Agreements between user and aligned model. Users can check when both they and the model succeeded or failed.
\end{minipage}
\begin{minipage}{2\columnwidth}
\centering
\includegraphics[width=0.6\textwidth]{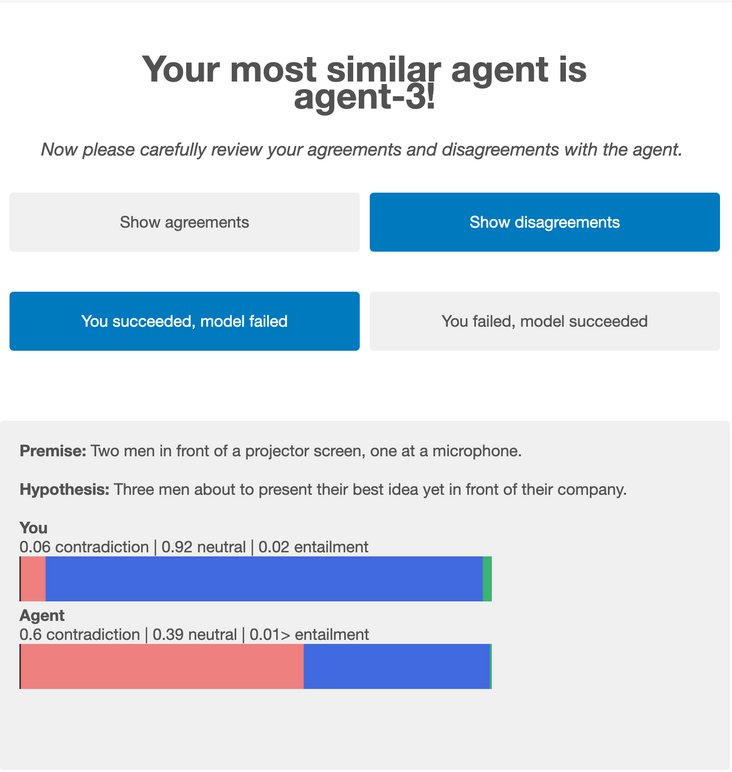}\\
(b) Disagreements between user and aligned model. Users can check when they succeeded but the model failed and vice versa.\\
\captionof{figure}{Agreements/disagreements between user and the aligned model.}
\end{minipage}

\clearpage

\section{Human Annotation Aggregation}
We compare our user study results with aggregated user labels. Specifically,  we aggregate 100 users' hard label answers and call the label distribution as \textbf{aggUSER}. We then calculate average user divergences in the user study \textbf{(avgUSER)}, as we did in the study, but this time with all users, and compare their divergences to the models. As seen in Table \ref{table:model-group-distances}, while mostly the order of divergences to models is preserved, for $\beta$ -KL, the LSTM changes place with Davinci. Therefore, we again emphasize the difference between the two methods averaging and aggregating: especially studying with low-level sample sizes.\\

\begin{minipage}{2\columnwidth}
\centering
\begin{tabular}{|p{0.18\linewidth}|p{0.18\linewidth}|p{0.18\linewidth}|p{0.18\linewidth}|p{0.18\linewidth}|}
\hline
\textbf{Agent} & \textbf{Ag \%} & \textbf{$\alpha$-KL} & \textbf{$\beta$-KL} & \textbf{JSD} \\
                & \textbf{(avg/agg)} & \textbf{(avg/agg)} & \textbf{(avg/agg)} & \textbf{(avg/agg)} \\
\hline
LSTM            & 0.64 / 0.74 & 0.73 / 0.47 & 1.48 / 0.64 & 0.25 / 0.25 \\
RoBERTa         & 0.68 / 0.82 & 0.56 / 0.30 & 1.23 / 0.47 & 0.24 / 0.20 \\
Davinci         & 0.51 / 0.54 & 2.76 / 2.61 & 1.33 / 1.06 & 0.38 / 0.42 \\
\hline
\multicolumn{5}{|c|}{\textbf{Accuracy (avgUSER-aggUSER)}} \\
\multicolumn{5}{|c|}{0.70 / 0.84} \\
\hline
\end{tabular}
\captionof{table}{Ag\% represents the percentage of questions where the model-group agrees on its \textit{hard} label. The table shows the model distances of the aggregated human-100 group and the average user in a user study. Except for $\beta$-KL, the order of models' divergences is preserved.}
\label{table:model-group-distances}
\end{minipage}
\end{document}